\documentclass[conference]{IEEEtran}
\IEEEoverridecommandlockouts
\usepackage{cite}
\usepackage{amsmath,amssymb,amsfonts}
\usepackage{graphicx}
\usepackage{textcomp}
\usepackage{xcolor}
\usepackage{algorithm}
\usepackage{algorithmic}

\usepackage[caption=false, font=footnotesize]{subfig}

\usepackage[capitalise]{cleveref}

\usepackage{flushend}

\def\BibTeX{{\rm B\kern-.05em{\sc i\kern-.025em b}\kern-.08em
    T\kern-.1667em\lower.7ex\hbox{E}\kern-.125emX}}
\begin{document}

\title{Chemical Property-Guided Neural Networks for Naphtha Composition Prediction
% \thanks{Identify applicable funding agency here. If none, delete this.}
}

\author{\IEEEauthorblockN{
                Chonghyo Joo\IEEEauthorrefmark{1}\IEEEauthorrefmark{2}\thanks{C. Joo and J. Kim - These authors contributed equally to this work.} 
                Jeongdong Kim\IEEEauthorrefmark{1},                
                Hyungtae Cho\IEEEauthorrefmark{2},
                Jaewon Lee\IEEEauthorrefmark{2},
                Sungho Suh\IEEEauthorrefmark{3}\IEEEauthorrefmark{4},
                Junghwan Kim\IEEEauthorrefmark{1}}
		\IEEEauthorblockA{\IEEEauthorrefmark{1}Department of Chemical and Biomolecular Engineering, Yonsei University, Seoul, Republic of Korea}
		\IEEEauthorblockA{\IEEEauthorrefmark{2}Green Materials \& Processes R\&D Group, Korea Institute of Industrial Technology, Ulsan, Republic of Korea}
		\IEEEauthorblockA{\IEEEauthorrefmark{3}Department of Computer Science, RPTU Kaiserslautern-Landau, Kaiserslautern, Germany}
        \IEEEauthorblockA{\IEEEauthorrefmark{4}German Research Center for Artificial Intelligence (DFKI), Kaiserslautern, Germany}
		Email: hyo156@yonsei.ac.kr, jdong9604@yonsei.ac.kr,  htcho@kitech.re.kr,\\ j.lee@kitech.re.kr, sungho.suh@dfki.de, kjh24@yonsei.ac.kr}
  
% \author{\IEEEauthorblockN{1\textsuperscript{st} Given Name Surname}
% \IEEEauthorblockA{\textit{dept. name of organization (of Aff.)} \\
% \textit{name of organization (of Aff.)}\\
% City, Country \\
% email address or ORCID}
% \and
% \IEEEauthorblockN{2\textsuperscript{nd} Given Name Surname}
% \IEEEauthorblockA{\textit{dept. name of organization (of Aff.)} \\
% \textit{name of organization (of Aff.)}\\
% City, Country \\
% email address or ORCID}
% \and
% \IEEEauthorblockN{3\textsuperscript{rd} Given Name Surname}
% \IEEEauthorblockA{\textit{dept. name of organization (of Aff.)} \\
% \textit{name of organization (of Aff.)}\\
% City, Country \\
% email address or ORCID}
% \and
% \IEEEauthorblockN{4\textsuperscript{th} Given Name Surname}
% \IEEEauthorblockA{\textit{dept. name of organization (of Aff.)} \\
% \textit{name of organization (of Aff.)}\\
% City, Country \\
% email address or ORCID}
% \and
% \IEEEauthorblockN{5\textsuperscript{th} Given Name Surname}
% \IEEEauthorblockA{\textit{dept. name of organization (of Aff.)} \\
% \textit{name of organization (of Aff.)}\\
% City, Country \\
% email address or ORCID}
% \and
% \IEEEauthorblockN{6\textsuperscript{th} Given Name Surname}
% \IEEEauthorblockA{\textit{dept. name of organization (of Aff.)} \\
% \textit{name of organization (of Aff.)}\\
% City, Country \\
% email address or ORCID}
% }

\maketitle

\begin{abstract}
The naphtha cracking process heavily relies on the composition of naphtha, which is a complex blend of different hydrocarbons. Predicting the naphtha composition accurately is crucial for efficiently controlling the cracking process and achieving maximum performance. Traditional methods, such as gas chromatography and true boiling curve, are not feasible due to the need for pilot-plant-scale experiments or cost constraints. In this paper, we propose a neural network framework that utilizes chemical property information to improve the performance of naphtha composition prediction. Our proposed framework comprises two parts: a Watson K factor estimation network and a naphtha composition prediction network. Both networks share a feature extraction network based on Convolutional Neural Network (CNN) architecture, while the output layers use Multi-Layer Perceptron (MLP) based networks to generate two different outputs - Watson K factor and naphtha composition. The naphtha composition is expressed in percentages, and its sum should be 100\%. To enhance the naphtha composition prediction, we utilize a distillation simulator to obtain the distillation curve from the naphtha composition, which is dependent on its chemical properties. By designing a loss function between the estimated and simulated Watson K factors, we improve the performance of both Watson K estimation and naphtha composition prediction. The experimental results show that our proposed framework can predict the naphtha composition accurately while reflecting real naphtha chemical properties. 
\end{abstract}


\section{Introduction}
\begin{figure}[!t]
    \centering
    \includegraphics[width=\columnwidth]{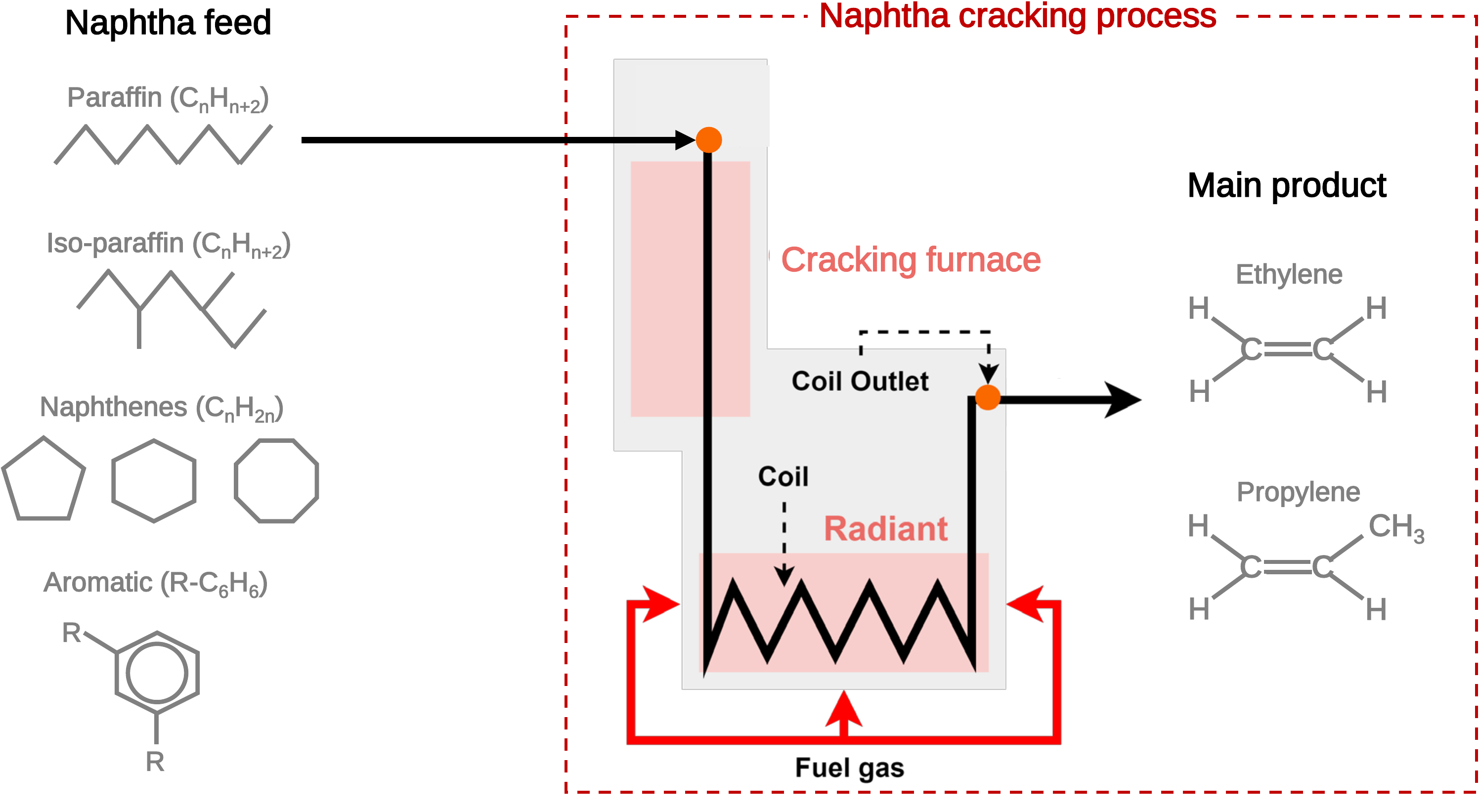}
    \caption{Overview of the naphtha cracking center (NCC) for ethylene and propylene production}
    \label{fig:ncc}
\end{figure}

Naphtha, a liquid hydrocarbon mixture derived from crude oil, serves as a crucial feedstock in Naphtha Cracking Centers (NCCs) for the production of core chemicals such as methane, ethylene, and propylene as shown in \cref{fig:ncc}. These chemicals are essential in the manufacturing of plastics, synthetic rubbers, and other materials widely used across various industries. The ratio of produced chemicals in NCCs is influenced by several control variables. In order to enhance economic value, it is imperative to manipulate these variables to optimize the production of target products in NCC. The yield of products depends on the two key variables: 1) operating conditions such as operating temperature and pressure, and 2) the naphtha composition such as paraffins and aromatics \cite{ren2019molecular}. The composition of the naphtha is significant because each component has a different activation energy required to be cracked into lighter hydrocarbons. In the past, the chemical industry has employed optimization and estimation techniques to increase the yield of the target products. The operating conditions have been optimized using various commercial simulation software \cite{van2010model, joo2000cracker,rezaeimanesh2022coupled,amghizar2017decision} and the detail naphtha composition has been estimated roughly \cite{riazi2005characterization,van2007molecular,dente1979detailed} for feedstock management. However, the traditional approaches rely heavily on mathematical equations and theoretical models coupled with case studies, making the optimization process time-consuming. Additionally, determining the precise composition of naphtha is challenging due to its complexity, despite its significant influence on product yield. Therefore, relying on mathematical equations or theoretical models is not feasible as naphtha’s detailed composition can only be estimated through various experiments, such as gas chromatography, which offers limited flexibility. 

Recent advances in machine learning and data analytics have provided new opportunities for the optimization of cracking furnace with high accuracy and efficiency. Kim et al. developed machine learning-based yield prediction model of cracking process \cite{kim2023multi}.The model was applied to find out the optimal operating conditions for increasing the product yield considering uncertainty. As a result, they found the optimal operating conditions according to different target products, ethylene and propylene. Plehiers et al. developed artificial neural networks (ANNs)-based model to predict a detailed characterization of naphtha and the flow rate of products \cite{plehiers2019artificial}. The detailed characterization of a naphtha is predicted from three boiling points and main five components of naphtha. The flow rate of products was estimated using the reconstructed naphtha and operating conditions. Although the accuracy of their model was lower than commercial simulation software, the required central processing unit time per reaction was in the order of seconds. Ma et al. \cite{ma2022molecular} suggested a neural network-based naphtha molecular reconstruction model. They improved the typical artificial neural network-based model by embedding physical information into the neural network. Mei et al. \cite{mei2017molecular} proposed a molecular-based Bayesian regression model for petroleum fractions. They assumed that each component of naphtha is independently and identically distributed. Bi and Qiu developed a novel naphtha molecular reconstruction process using a genetic algorithm and particle swarm optimization \cite{bi2019intelligent}. The proposed model was designed to provide an accurate composition for local refineries. They defined probability density functions to apply an optimization algorithm. As a result, they found that the proposed method reconstructed naphtha composition more accurately than other methods.  

Although considerable efforts have been made in previous studies to optimize naphtha cracking processes, most of these studies have primarily focused on optimizing operating conditions. A few studies have attempted to reconstruct naphtha composition in detail and demonstrated promising results; however, they often relied on several assumptions or neglected other important indicators, such as the Watson K factor, which encapsulates the chemical property information of naphtha. This study aims to address these limitations by comprehensively considering both detailed composition data and chemical properties for the optimization of naphtha cracking processes.

%To address this challenge, we propose a chemical property-guided neural network framework for accurate naphtha composition prediction using real-world naphtha data. Our proposed framework utilizes chemical property information to improve the performance of naphtha composition prediction. Specifically, our framework comprises two parts: a Watson K factor estimation network and a naphtha composition prediction network. The Watson K factor is a measure of the hydrocarbon mixture's normal paraffin content, which plays a crucial role in accurately predicting the naphtha composition.

%Naphtha cracking process is a crucial step in the petrochemical industry, as the process is used to convert naphtha into a range of useful chemicals and petrochemicals as shown in \cref{fig:ncc}. Naphtha, a liquid hydrocarbon mixture derived from crude oil, is used as feedstock in the naphtha cracking center (NCC) to produce joint products including methane, ethylene, and propylene. However, since the NCC is a joint product process, it cannot exclusively produce the target products. Therefore, it is necessary to control some variables to increase the production of the target products in NCC.

To address this challenge, we propose a chemical property-guided neural network framework for accurate naphtha composition prediction using real-world naphtha data. Our proposed framework utilizes chemical property information to improve the performance of naphtha composition prediction. Specifically, our framework comprises two parts: a Watson K factor estimation network and a naphtha composition prediction network. The Watson K factor is a measure of the hydrocarbon mixture's normal paraffin content, which plays a crucial role in predicting the naphtha composition accurately The Watson K factor estimation network and naphtha composition prediction network share a feature extraction network based on the Convolutional Neural Network (CNN) architecture, while the output layers use Multi-Layer Perceptron (MLP) based networks to generate two different outputs - Watson K factor and naphtha composition. To enhance the naphtha composition prediction, we utilize a distillation simulator to obtain the distillation curve from the naphtha composition, which is dependent on its chemical properties. By designing a loss function between the estimated and simulated Watson K factors, we improve the performance of both Watson K factor estimation and naphtha composition prediction. The experimental results show that our proposed framework can predict the naphtha composition accurately and can be applied to chemical engineering processes.

%%To further enhance the naphtha composition prediction, we utilize a distillation simulator to obtain the distillation curve from the naphtha composition, which is dependent on its chemical properties. By designing a loss function between the estimated and simulated Watson K values, we improve the performance of both Watson K estimation and naphtha composition prediction. The experimental results show that our proposed framework can predict the naphtha composition accurately and can be applied to chemical engineering processes.

\begin{figure*}[!t]
    \centering
    \includegraphics[width=\linewidth]{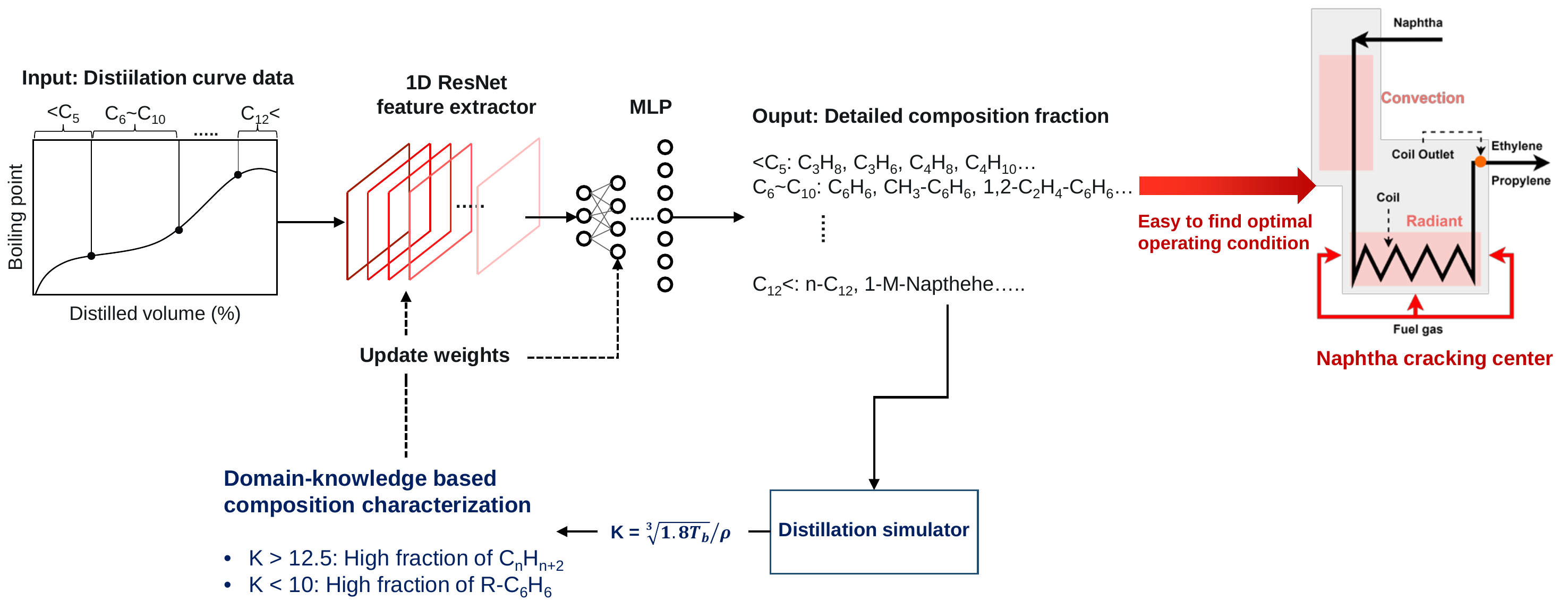}
    \caption{Schematic illustration of proposed chemical property-guided model}
    \label{fig:method_overview}
\end{figure*}

The main contributions of our study can be summarized as follows:
\begin{itemize}
    \item This is the first study to predict detailed naphtha composition from a distillation curve using deep neural networks. To predict the naphtha composition in detail, we propose a novel naphtha composition prediction model.
    \item To improve the naphtha composition prediction model, we propose a chemical property-guided neural network framework by using the Watson K factor.
    \item The proposed framework is validated with real industrial naphtha data, collected from a real naphtha cracking process. Thus, this framework is expected to be more applicable to real-world plants. 
\end{itemize}

The remainder of this paper is organized as follows. \cref{sec:proposedmthod} provides an overview of the network architecture and training process. \cref{sec:experiments} presents our experimental results and analysis. Finally, in \cref{sec:conclusion}, we conclude our work and give an insight into future work.

\section{Methodology}
\label{sec:proposedmthod}
The goal of the proposed framework is to predict the detailed composition of naphtha with high accuracy to quickly respond to the cracking process. By utilizing the predicted composition information, it is possible to optimize the reaction temperature for maximizing product yields using thermodynamic reaction kinetics equations. In the industrial fields, domain experts roughly estimate the composition of general types components of naphtha by analyzing the corresponding distillation curve. To predict the naphtha composition accurately, we propose the chemical property-guided model, as shown in \cref{fig:method_overview}. The proposed model takes the distillation curve data, which is determined by the physicochemical interaction between detailed composition, as input. The proposed model consists of a feature extractor motivated from ResNet \cite{he2016deep} and MLP-based networks that predict the weight fraction of the detailed chemical composition. To guide the training procedure and improve the naphtha composition performance, we incorporate the Watson K factor, which characterizes different types of chemicals, including paraffin, aromatic, and naphthene-type chemicals. The Watson K factor can be calculated by a commercial chemical simulator and a traditional physics equation. In the training procedure, Thus, the additional loss term of K value updates the network to predict the naphtha having realistic chemical properties, not merely having high prediction accuracy.

\subsection{Problem definition} %problem 정의가 필요 input, output 얘기, 
Predicting the detailed naphtha composition in the industry is a challenging task that often requires multiple experiments. The conventional approaches have limited flexibility because they rely on a restricted set of commercial indices. In addition, recently proposed data-driven models require certain assumptions, which may hinder their application in real-world industrial settings. This study proposes a novel method for naphtha prediction using real plant data that considers a commercial chemical indicator to overcome these challenges. The proposed model utilizes the distillation curve of naphtha $X = [x_1, ..., x_{n_d}]$ with $n_d$ boiling points of the distillation curve. Consequently, to compute the distillation curve of naphtha as a composition-dependent variable using the simulator, this research employed data preprocessing techniques to transform the flow units (ton/hour) data of naphtha into weight percent (wt\%) units data. This conversion was necessary to ensure that the total components of the naphtha add up to 100 wt\%. 

The Watson K factor $k$ as input variables and the corresponding naphtha composition $C = [c_1, ..., c_{n_c}]$ with $n_c$ compositions as output variables. The Watson K factor, which classifies oil by boiling points and density, is calculated using 
\cref{eq:k_factor} \cite{watson1935characterization,daubert1997american,speight2014chemistry}:
\begin{equation}
\label{eq:k_factor}
    k = \frac{\sqrt[3]{1.8\tau b}}{\rho}
\end{equation}
where $\tau$ is the reference temperature, $b$ is a coefficient, and $\rho$ is the oil density. 

% \subsection{Data-driven modeling}
%For chemical property-guided neural network modeling, Watson K simulator and neural network were applied. \cref{fig:feed_back_loop} shows the schematic illustration of the proposed feedback loop model structure. The networks were designed to generate two different outputs - naphtha composition and Watson K factor. Naphtha composition prediction is the final goal of the proposed model, and Watson K factor prediction is applied to update the weights in the networks which can enhance the naphtha composition prediction. Therefore, two loss functions were designed for feedback loop modeling and used to update the weights of neural networks alternately: the first loss function is a mean squared error (MSE) between the calculated and predicted naphtha composition, and the second one is an MSE between the calculated and predicted Watson K factors.

\subsection{Naphtha Composition Prediction Model}
% \subsubsection{CNN-based model}
%As shown in \cref{fig:feed_back_loop}, naphtha composition data converts to distillation curve data to use the distillation curve data as inputs of the model. The reason for use of distillation curve as inputs is that normally distillation curve not only depends on the naphtha composition, but also has been widely used to indicate the characteristics of naphtha.  

\begin{figure}[!t]
    \centering
    \includegraphics[width=\columnwidth]{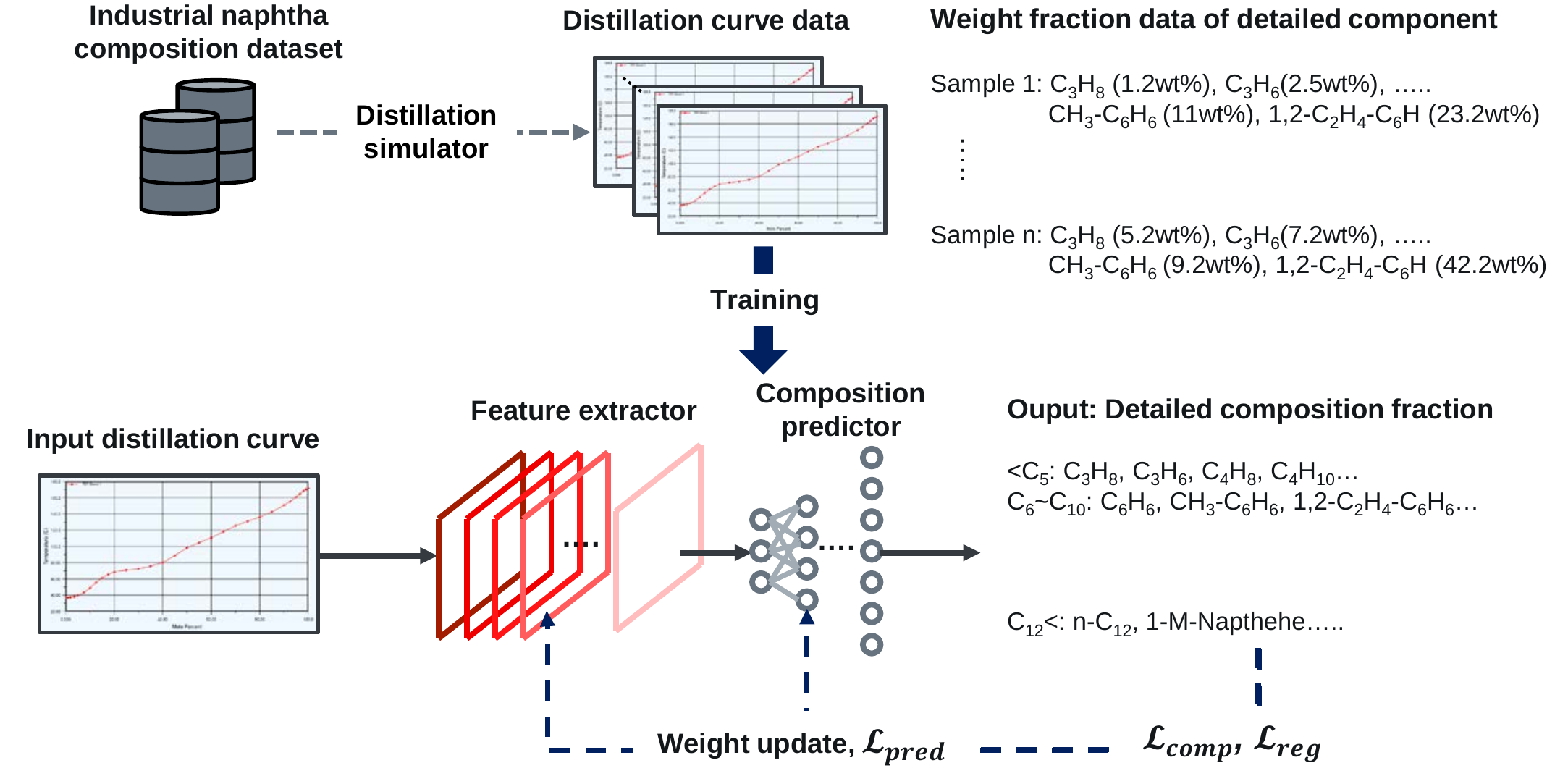}
    \caption{Training procedure and loss of the proposed naphtha composition prediction model}
    \label{fig:base_line}
\end{figure}

\begin{figure}[!t]
    \centering
    \includegraphics[width=\columnwidth]{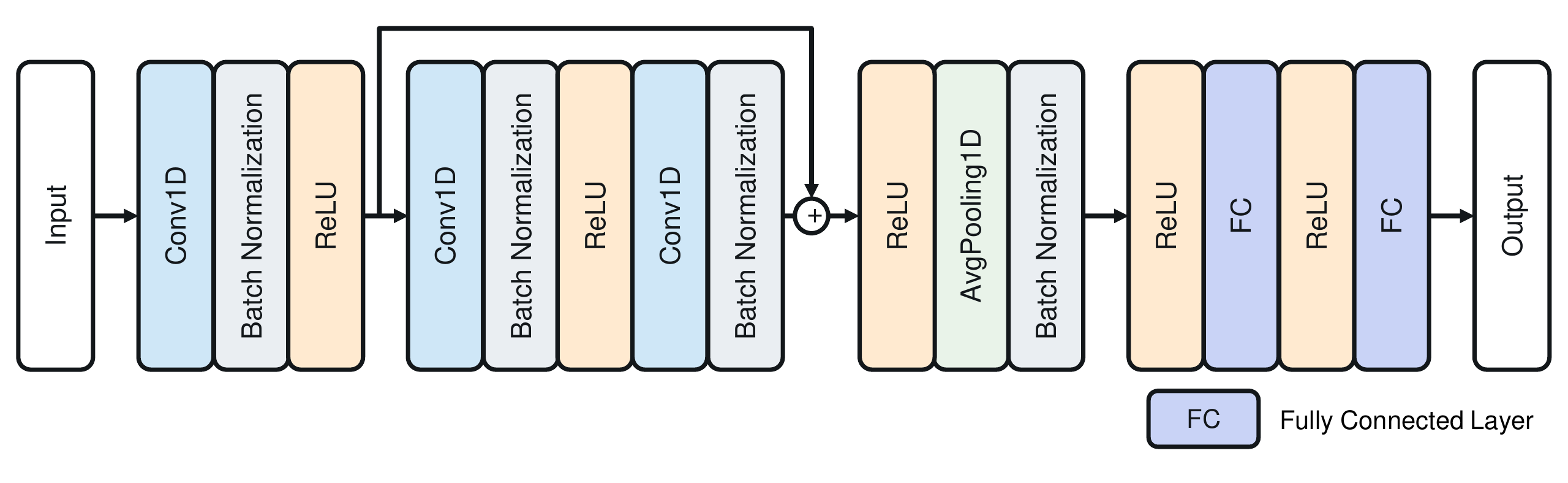}
    \caption{Detailed network structure of the feature extractor}
    \label{fig:resnet}
\end{figure}

The naphtha composition prediction model is illustrated in \cref{fig:base_line}. In the chemical industry, naphtha composition is typically estimated using a distillation curve and expert knowledge. To incorporate this knowledge into the model, we designed a feature extractor $f(\cdot)$ that is motivated from ResNet with 1D operations \cite{hong2020holmes}, as shown in \cref{fig:resnet}, an MLP-based composition estimation network $e(\cdot)$. The input for naphtha composition prediction is the distillation curve obtained by using ASPEN Hysys software \cite{haydary2019chemical,santos2021astm}. The ground truth of the naphtha composition $C$ is given in the dataset, and we preprocess the real naphtha compositions to match the sum of the compositions equal to 100 wt\%. The proposed naphtha composition prediction model is trained to predict the detailed naphtha components from the input distillation curve data with the preprocessed naphtha composition ground truth. 


To train the networks, we formulate two loss functions. The first loss function calculates the difference between the predicted compositions and the ground truth of real compositions. The second loss function regularizes the sum of the predicted compositions to be equal to 100 wt\%. 

% The CNN-based model is trained to predict the detailed naphtha components from the distillation curve data, consistent with the expert estimations of naphtha
% Here, the loss functions are described in \cref{eq:loss1,eq:loss2} for updating weights of the CNN-based model. \cref{eq:loss1}calculates the difference between the predicted and real compositions, while the one while \cref{eq:loss2} ensures that the sum of predicted compositions is equal to 100 weight percentage (wt\%). Thus, By minimizing the loss between the predicted and true values, the model can be trained to accurately predict compositions while maintaining their total sum as 100 wt\%.

\begin{equation}
\label{eq:loss1}
    \mathcal{L}_{comp} = \left\lVert C - e(f(X)) \right\rVert_2^2
\end{equation}
\begin{equation}
\label{eq:loss2}
    \mathcal{L}_{res} = \left\lVert 100 - \sum^{n_c}_j \hat{c}_j \right\rVert_2^2
\end{equation}
where $\hat{C}$ is the output of the model $e(f(X))$, $n_c$ is the number of components in naphtha, the objective loss $\mathcal{L}_{pred}$ is expressed as the weighted sum of two loss functions. The weights $\lambda_{comp}$ and $\lambda_{res}$ control the relative importance of different loss terms.
\begin{equation}
\label{eq:composition}
    \mathcal{L}_{pred} = \lambda_{comp} \mathcal{L}_{comp} + \lambda_{res} \mathcal{L}_{res}
\end{equation}

\subsection{Chemical Property-guided Neural Network Framework with Watson K Factor}
% \subsubsection{Chemical property-guided model}

\begin{figure}[!t]
    \centering
    \includegraphics[width=\columnwidth]{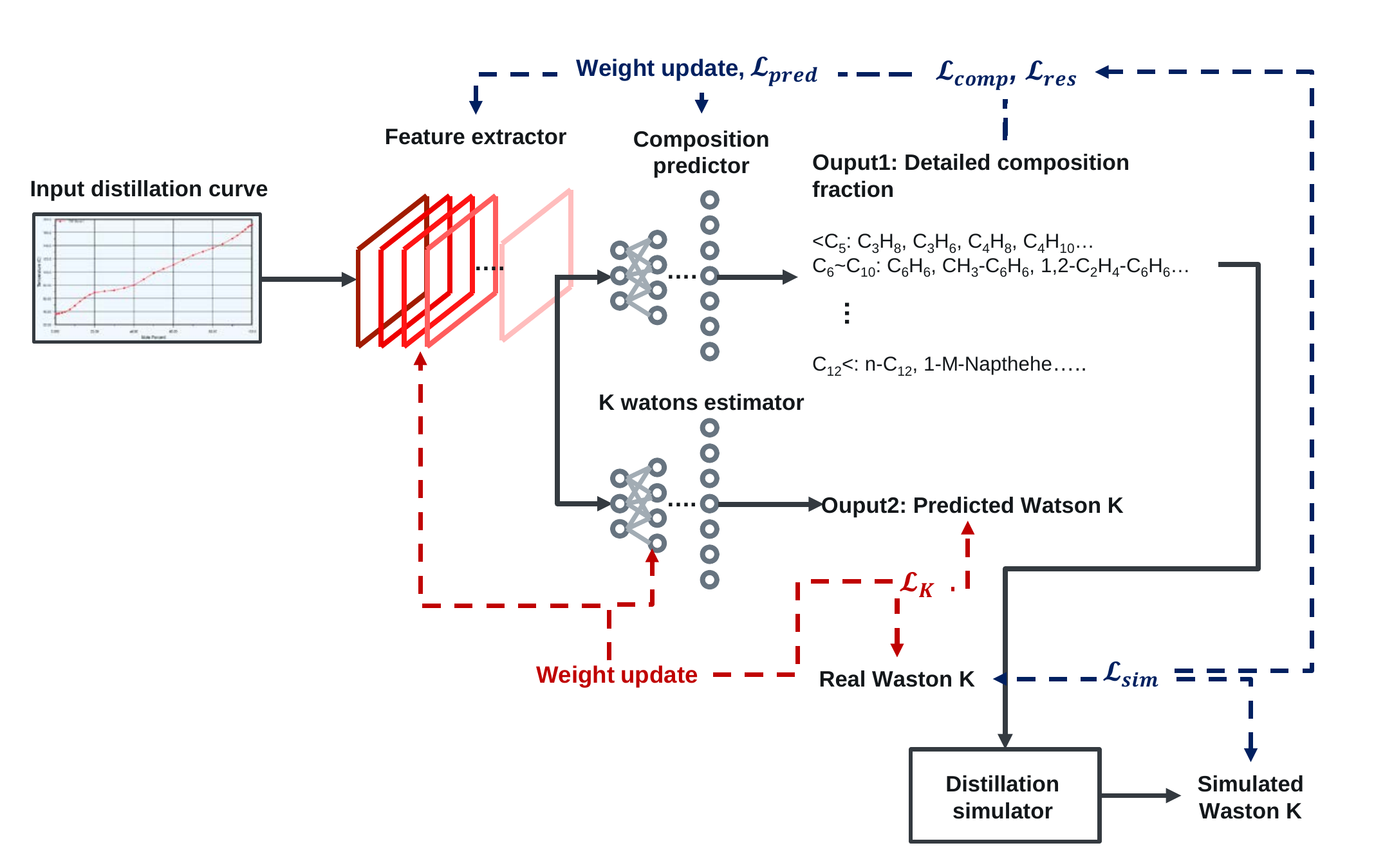}
    \caption{Overview of the training procedure and loss of the proposed chemical property-guided framework}
    \label{fig:feed_back_loop}
\end{figure}

To improve the naphtha composition prediction and incorporate the chemical property, we propose the chemical property-guided neural network model using the Watson K factor, a chemical property indicator of naphtha to guide the prediction of the naphtha composition. As shown in \cref{fig:feed_back_loop}, our proposed framework consists of three networks: a feature extractor $f(\cdot)$, an MLP-based composition estimation network $e(\cdot)$, and an MLP-based Watson K factor estimation network $w(\cdot)$. The weights of the feature extractor are shared between the composition estimation and the Watson K factor estimation networks. The composition estimation network is trained by minimizing the composition prediction loss in \cref{eq:composition} and an additional loss function that measures the discrepancy between the ground truth of the Watson K factor, calculated from the ground truth of the naphtha compositions, and the simulated Watson K factor value from the predicted naphtha composition.

\begin{equation}
    \label{eq:sim}
    \mathcal{L}_{Sim} = \left\lVert k - \hat{k}_{sim} \right\rVert ^2_2
\end{equation}
\begin{equation}
\label{eq:composition2}
    \mathcal{L}_{pred} = \lambda_{comp} \mathcal{L}_{comp} + \lambda_{res} \mathcal{L}_{res} + \lambda_{sim} \mathcal{L}_{sim}        
\end{equation}
where $\hat{k}_{sim}$ is the simulated Watson K factor of the predicted composition $e(f(X))$ by the distillation simulator and $\lambda_{sim}$ controls the relative importance of the simulated Watson K discrepancy loss.
On the other hand, the Watson K factor estimation network is trained by minimizing the discrepancy between the predicted Watson K factor and the ground truth of the Watson K factor.
\begin{equation}
\label{eq:watson}
    \mathcal{L}_{K} = \left\lVert k - w(f(X)) \right\rVert ^2_2
\end{equation}
This loss function enables the Watson K factor estimation network to estimate the Watson K factor without the simulation and the Watson K factor equation \cref{eq:k_factor}.
% The chemical property-guided model differs from the CNN-based model in that it incorporates Watson K, one of the chemical property indicators of naphtha. The proposed model utilizes an extra loss function \cref{eq:loss5} that measures the discrepancy between the predicted and actual Watson K factors. This loss function considers the total loss between real Watson K factors and non-real Watson K factors including simulated and predicted values. Thus, the loss function \cref{eq:loss5} enables the model to consider both the distillation curve and the chemical property of naphtha, which should result in more precise predictions of naphtha composition in detail. 
% \begin{equation}
% \label{eq:loss3}
%     \mathcal{L}_{PredK} = \frac{1}{n}\sum_{i=1}^{n} (k_i - \hat{k}_{pred})^{2}
% \end{equation}
% \begin{equation}
% \label{eq:loss4}
%     \mathcal{L}_{SimK} = \frac{1}{n}\sum_{i=1}^{n} (k_i - \hat{k}_{sim})^{2}
% \end{equation}

% \begin{equation}
% \label{eq:loss5}
%     \mathcal{L}_K = \beta_1 \mathcal{L}_{PredK} + \beta_2 \mathcal{L}_{SimK}
% \end{equation}
% where $K_i$ is the real Watson K factor of composition $i$, $\hat{k}_{pred}$ is the predicted Watson K factor of composition $i$ by the neural network-based model, and $\hat{k}_{sim}$ is the simulated Watson K factor of composition $i$ by the distillation simulator. 
To train the different networks, we cross-train the networks by minimizing the two different loss functions, respectively.
The detailed training procedure for the chemical property-guided neural network is demonstrated in \cref{alg:training}.
\begin{algorithm}[h]
		\caption{Training procedure for chemical property-guided neural networks $f$, $e$, and $w$. All experiments in the paper used the default values $m=8$, $\eta=0.01$, $n_{Watson}=2$.}
		\begin{algorithmic}[1]
			\REQUIRE Batch size $m$ and Adam hyperparameters $\eta$
			\STATE \textbf{Input:} Sets of distillation curves $x \in X$, real naphtha compositions $c \in C$, and Watson K factors $k \in K$
            \STATE \textbf{Output:} predicted composition $\hat{C}$ and the predicted Watson K factor $\hat{k}$ 
            \WHILE{$\theta_{f}, \theta_e, \theta_w$ have not converged}
		      \STATE Sample a mini-batch $(x, c, k)$ from the distillation curve data $X$, real naphtha compositions $C$, and the corresponding Watson K factor $K$.
                \STATE $\theta_f \gets \theta_f - \eta \nabla_{\theta_f} \mathcal{L}_{pred}(x, c, k;\theta_f),$
                \STATE $\theta_e \gets \theta_e - \eta \nabla_{\theta_e} \mathcal{L}_{pred}(x, c, k;\theta_e),$ \hfill$\triangleright$\cref{eq:composition2}
                \STATE Sample a mini-batch $(x, k)$ from the distillation curve data $X$, and the corresponding Watson K factor $K$.
                \STATE $\theta_f \gets \theta_f - \eta \nabla_{\theta_f} \mathcal{L}_{K}(x, k;\theta_f),$
                \STATE $\theta_w \gets \theta_w - \eta \nabla_{\theta_w} \mathcal{L}_{K}(x, k;\theta_w),$ \hfill$\triangleright$\cref{eq:watson}
            \ENDWHILE
		\end{algorithmic}
        \label{alg:training}
\end{algorithm}

\section{Experimental Results}
\label{sec:experiments}

%\subsubsection{Model evaluation}

\subsection{Dataset}
This study utilized actual naphtha data obtained from a Naphtha Catalytic Cracking (NCC) plant located in Korea. A total of 254 naphtha samples were gathered and subjected to chemical experiments to determine their detailed composition. The analysis revealed that the naphtha samples consisted of 25 primary components, including normal paraffins, isoparaffins, aromatics, and naphthene, based on their distinct carbon numbers. 

The distillation curve represents the boiling points of a liquid mixture as a function of their relative concentration. It shows the percentage of the liquid that boils at or below a given temperature and is commonly used  to characterize and analyze the composition of the oil. Here, the distillation curve data was collected from the simulator and 30 boiling points of the curve data, which ranges from 348 to 788 $K$, were applied to predict the naphtha composition. 

The dataset consisting of 254 samples including naphtha composition and distillation curve data is divided into two sets, one for training the model which contains 80\% of the data, and the other for testing, which contains the remaining 20\%. To avoid overfitting, K-fold cross-validation is performed in each epoch, where the training and validation datasets are randomly partitioned into 5 equal-sized folds.

\subsection{Implementation Details}
All models in this paper are trained using the Adam optimizer with a mini-batch size of 8. The learning rate hyperparameter was set to 0.01, and the initial decay rates for the first and second moments of the gradients were set to 0.9 and 0.99, respectively. Additionally, during the training of the Chemical property-guided model, weight parameters $\lambda_{comp}$, $\lambda_{res}$, and $\lambda_{sim}$ were set to 0.1, 0.001, and 1, respectively. 
Based on experiment studies. the optimal weights were chosen to ensure that each loss term contributed equally to the training loss. Once the training was completed, the prediction accuracy of both models was compared on the test dataset using the following prediction metrics:mean-absolute error ($MAE$), mean-squared error ($MSE$), and $R^{2}$. 

\begin{equation}
\label{eq:mae}
    MAE = \frac{1}{n}\sum_{i=1}^{n} \lvert y_i - \hat{y_i} \rvert
\end{equation}

\begin{equation}
\label{eq:mse}
    MSE = \frac{1}{n}\sum_{i=1}^{n} (y_i - \hat{y_i})^{2}
\end{equation}

\begin{equation}
\label{eq:r2}
    R^2 = 1 - \frac{\sum_{i=1}^{n} (y_i - \hat{y_i})^{2}}{\sum_{i=1}^{n} (y_i - \bar{y_i})^{2}}
\end{equation}
where $y_i$ is the true value, $\hat{y_i}$ is the predicted value, and $\bar{y_i}$ is the average value of the true values.

\begin{figure}[!t]
    \centering
    \subfloat[]{\includegraphics[width=0.5\columnwidth]{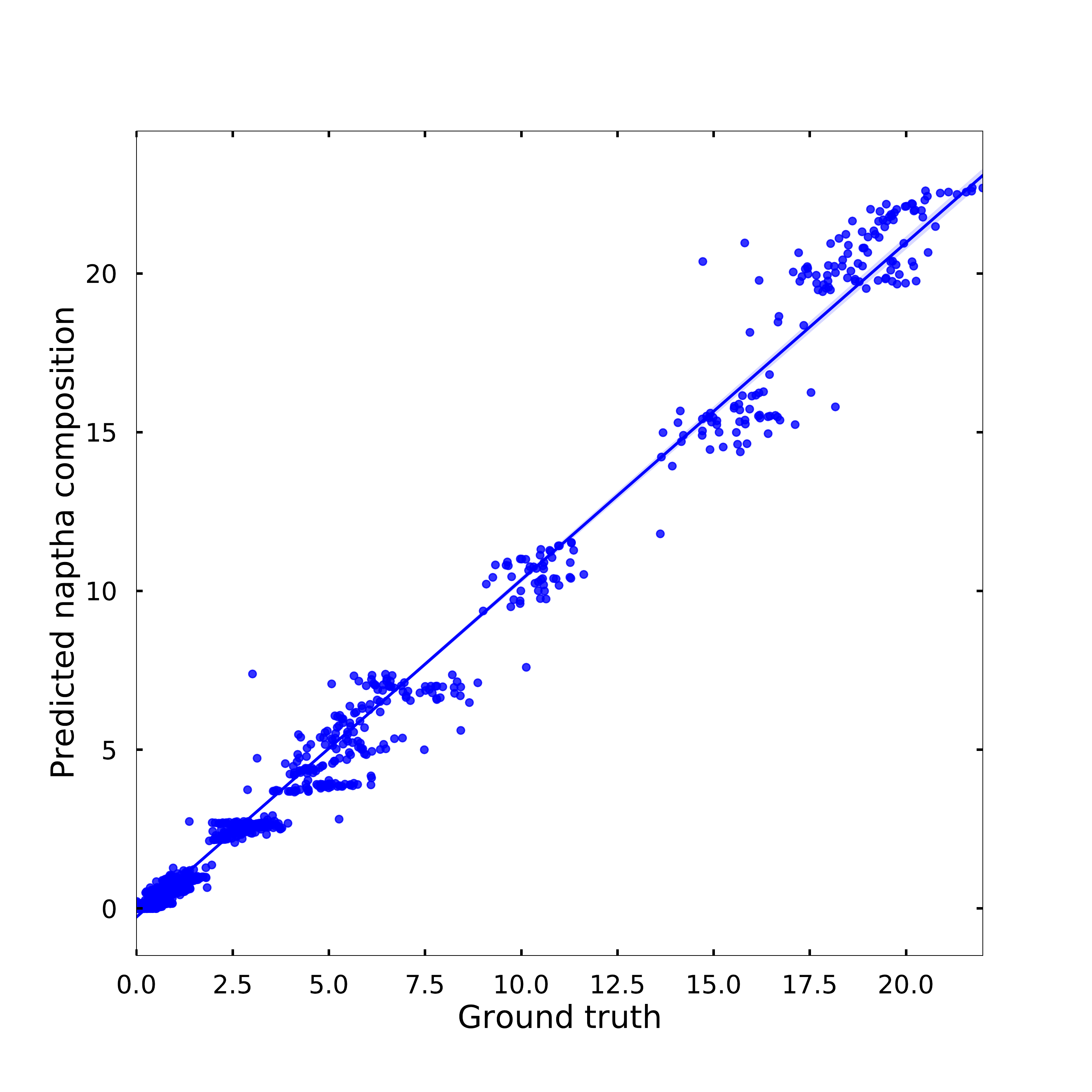}}
    % \subfloat[]{\includegraphics[width=0.5\columnwidth]{Figure/Parityplot_CNN_based.png}}
    % \hfill
    \subfloat[]{\includegraphics[width=0.5\columnwidth]{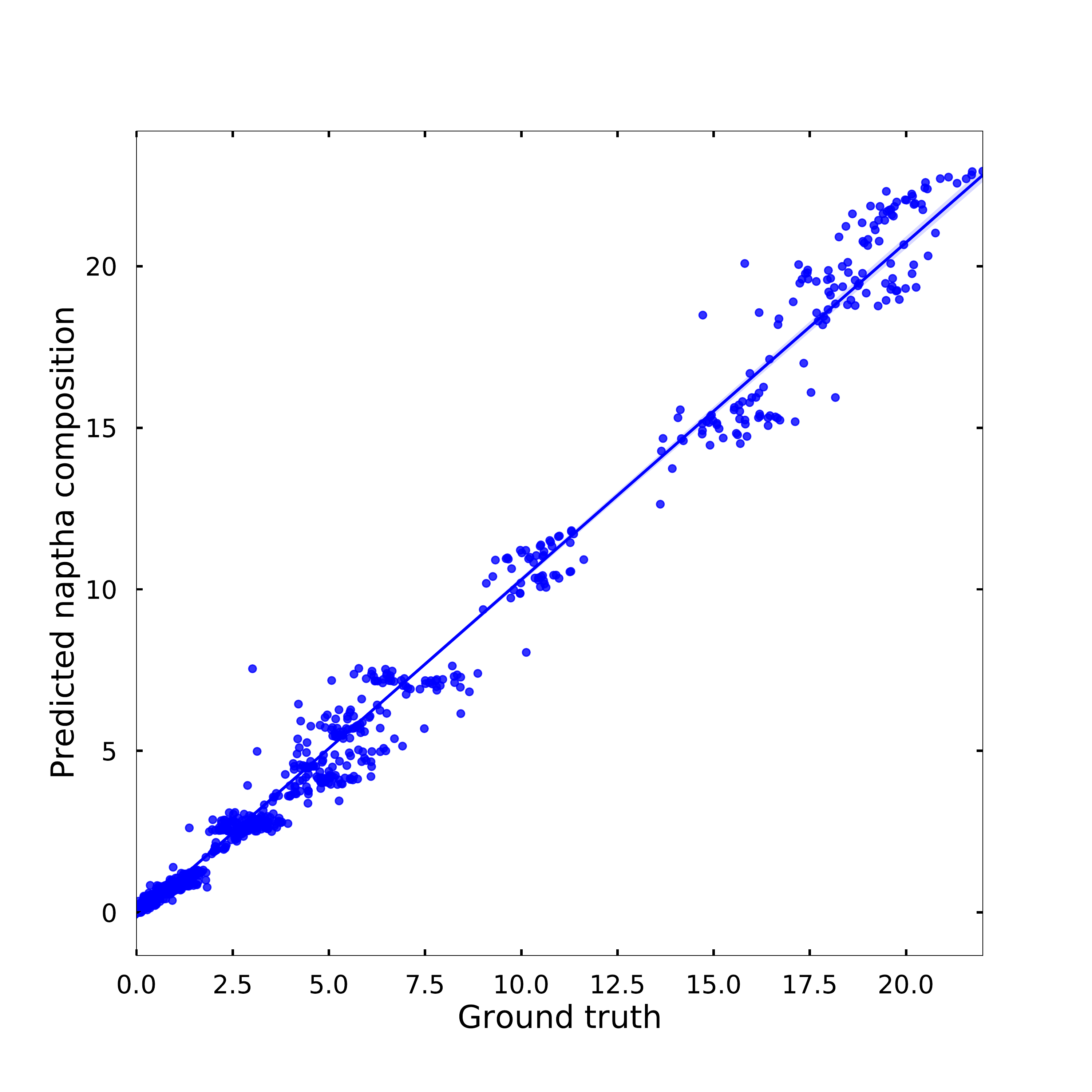}}
    % \subfloat[]{\includegraphics[width=0.5\columnwidth]{Figure/Parityplot_chemical_guided.png}}
    \caption{Parity plots for naphtha composition of different model: (a) CNN-based model; (b) Chemical property-guided model} 
    \label{fig:Parity plots}
\end{figure}

\begin{figure}[!t]
    \centering
    \subfloat[]{\includegraphics[width=0.5\columnwidth]{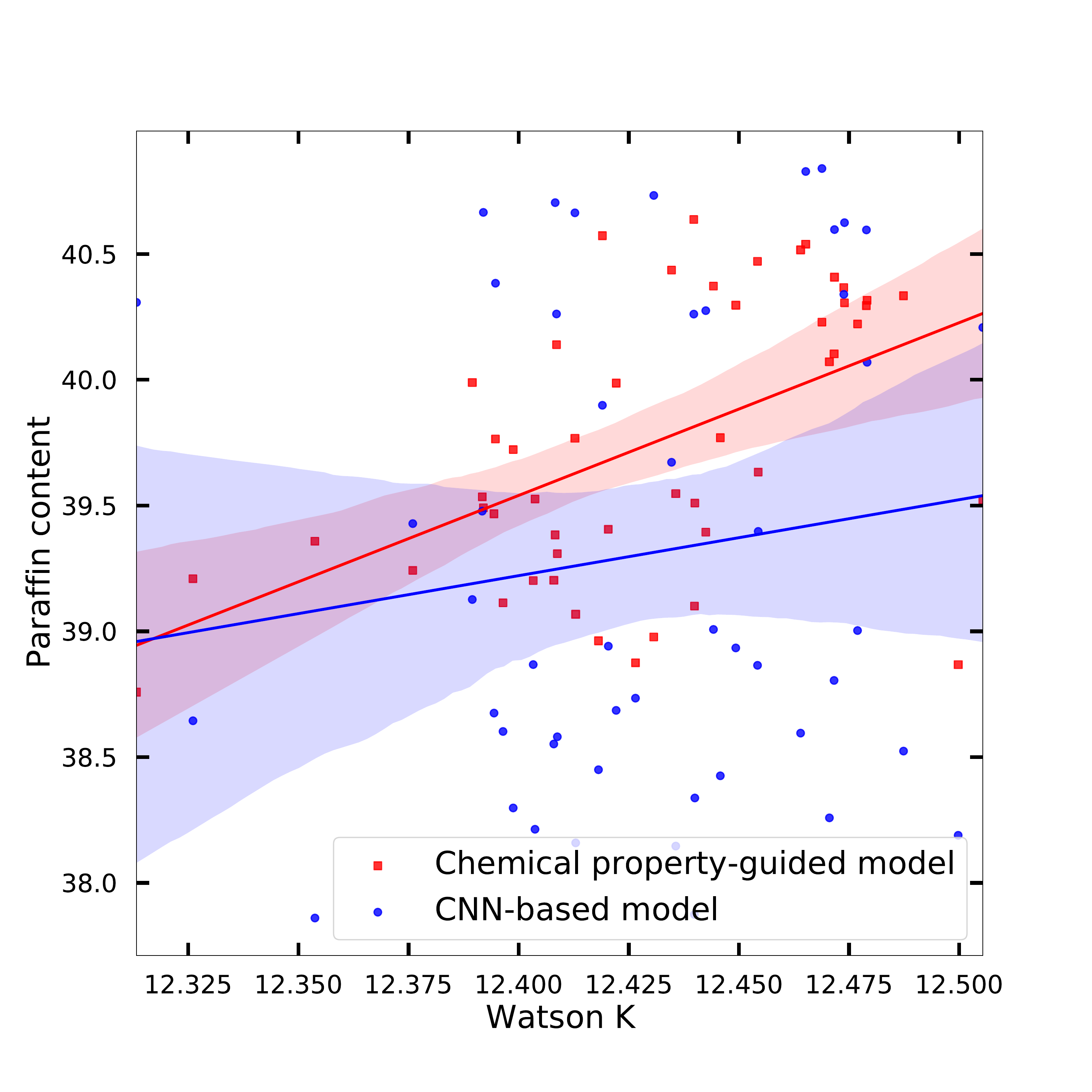}}
    % \subfloat[]{\includegraphics[width=0.5\columnwidth]{Figure/WatsonKvsParaffin.png}}
    % \hfill
    \subfloat[]{\includegraphics[width=0.5\columnwidth]{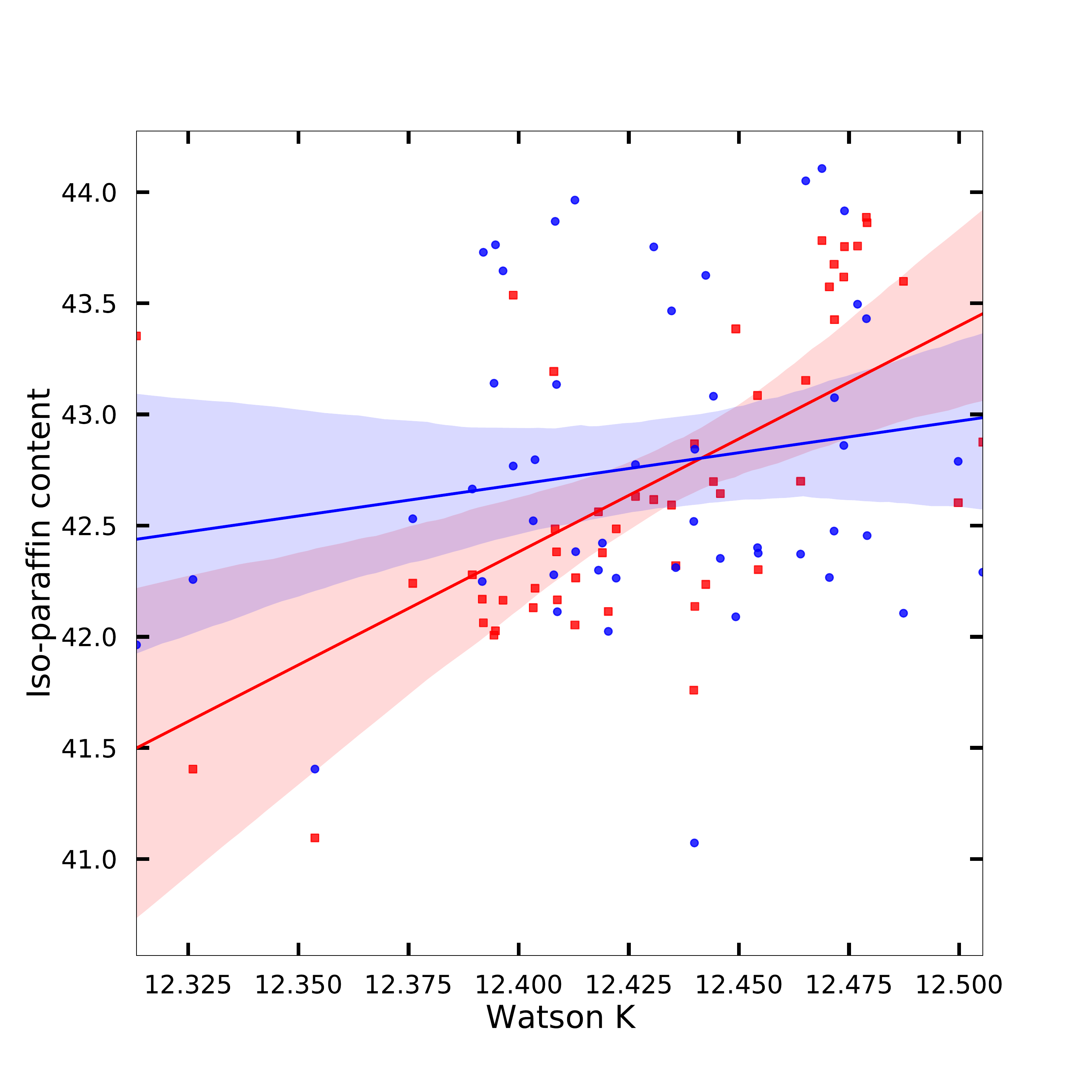}}
    % \subfloat[]{\includegraphics[width=0.5\columnwidth]{Figure/WatsonKvsIso-paraffin.png}}
    \hfill
    \subfloat[]  
    {\includegraphics[width=0.5\columnwidth]{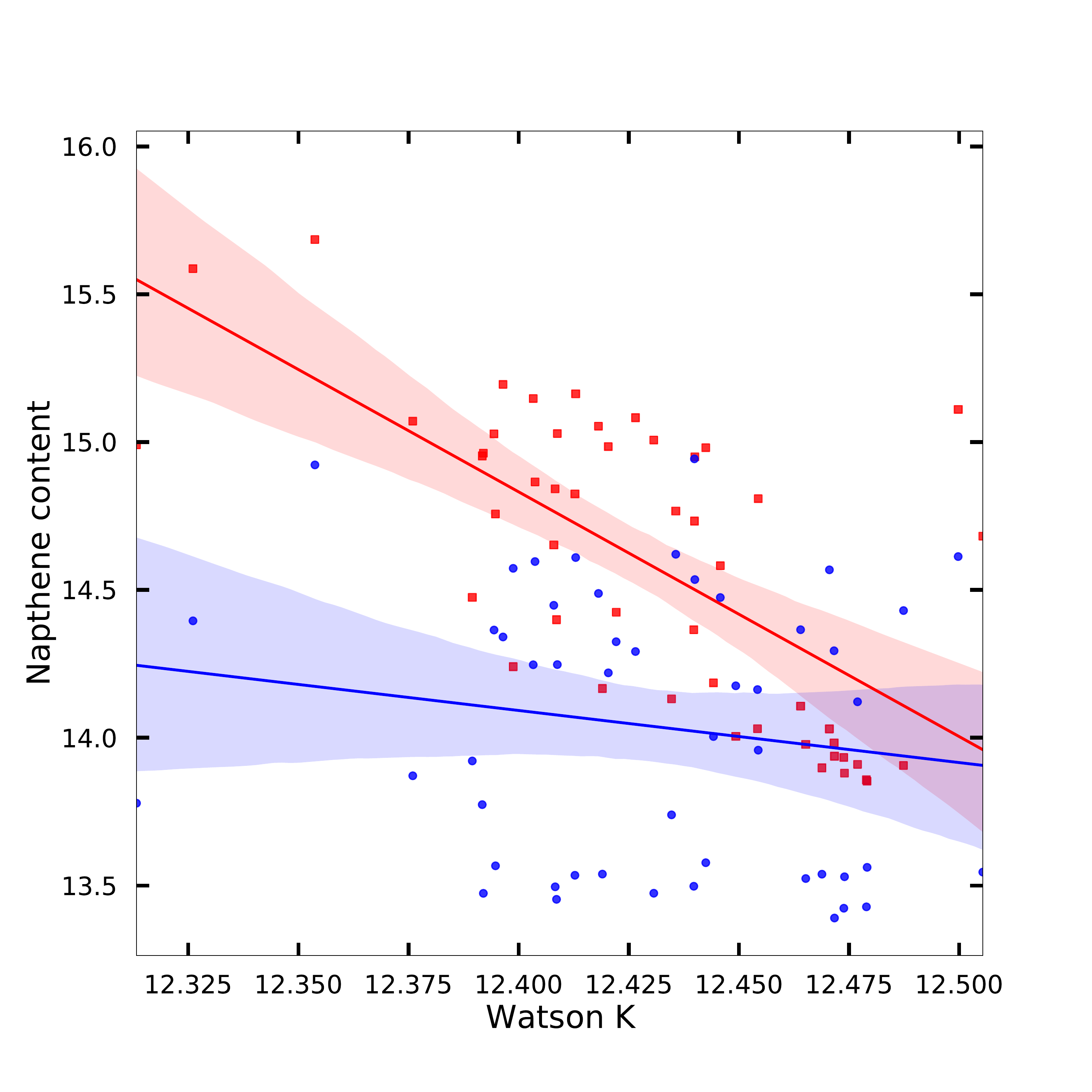}}
    % {\includegraphics[width=0.5\columnwidth]{Figure/WatsonKvsNaphthene.png}}
    \subfloat[]{\includegraphics[width=0.5\columnwidth]{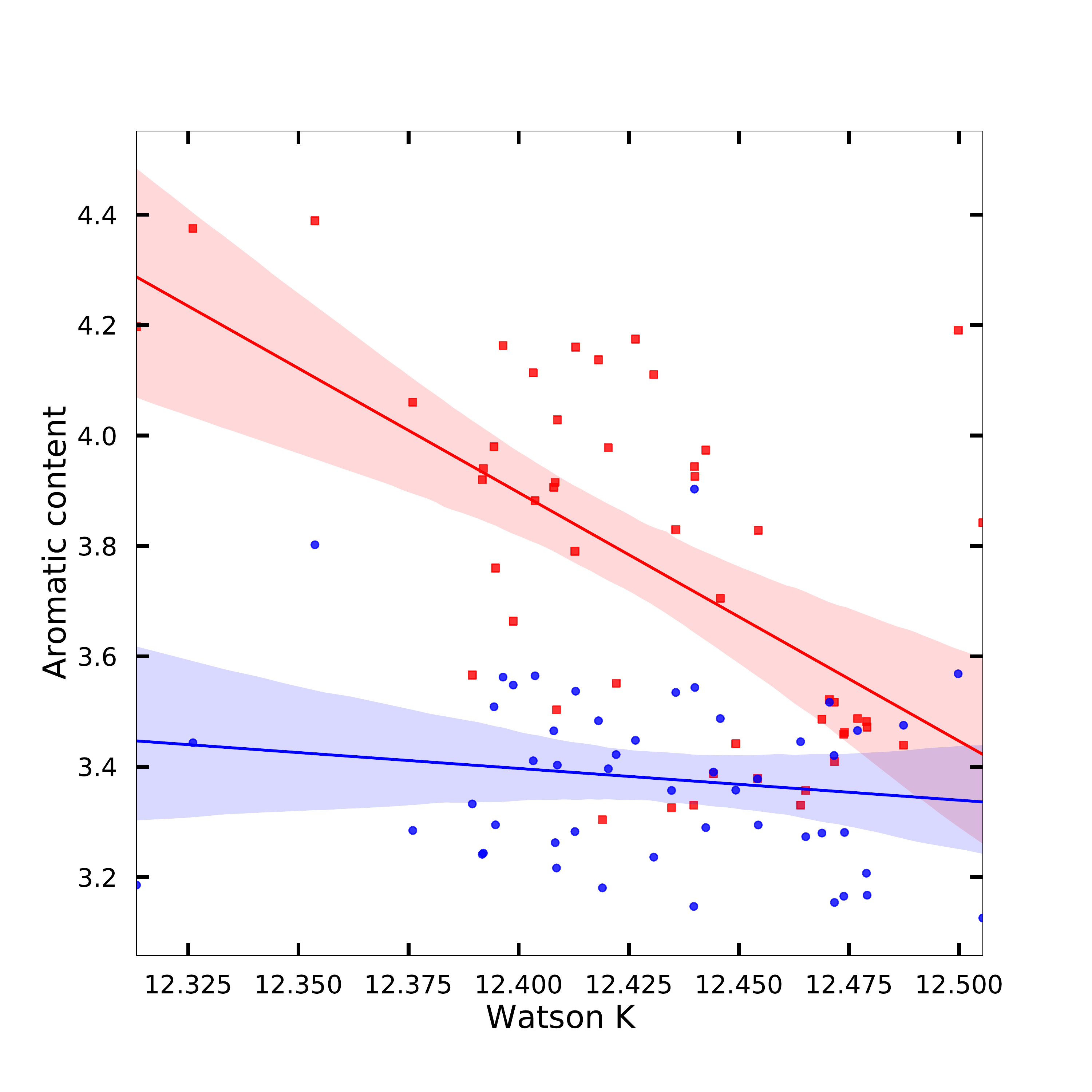}}
    % \subfloat[]{\includegraphics[width=0.5\columnwidth]{Figure/WatsonKvsAromatic.png}}
    \caption{Predicted general type components in different Watson K value: (a) Paraffin; (b) Iso-paraffin; (c) Naphthene ; (d) Aromatic content compound} 
    \label{fig:Watson vs P N A}
\end{figure}

\subsection{Results and Discussion}

For the test dataset, \cref{fig:Parity plots} shows the parity plots of the proposed models, CNN-based and chemical property-guided models. In the graph, the x- and y-axis indicate the true and predicted wt\% of each component, respectively. The data points are distributed near the line, y is equal to x, which implies that both models show high prediction performance for naphtha composition. Furthermore, \cref{table:train_result} lists the averaged prediction performance of the proposed models in 5-fold cross-validation. The CNN-based model for predicting naphtha composition had a good performance, with the averaged $MAE$ of 0.0213, $MSE$ of 0.0485, and $R^2$ of 0.95 after being validated with a five-fold cross-validation. However, the standard deviations of $MSE$ and $R^2$ were 0.029 indicating some variations among the folds. On the other hand, the chemical property-guided neural networks achieved an average $MAE$ of 0.0165, $MSE$ of 0.0327, and $R^2$ of 0.965 indicating that the chemical property-guided model not only performed for naphtha composition prediction but also reduced the standard deviations of $MSE$ and $R^2$ to 0.020 and 0.021, respectively. These results show that the chemical property-guided model has improved the performance of the naphtha prediction model by additional training for Watson K. 

In naphtha, according to the K factor, the general types of compositions follow specific behavior of weight fraction due to its physicochemical interaction. Commonly, in high Watson K value, the paraffin and iso-paraffin-type composition has a large weight fraction, aromatic and naphthene-type composition has a lower weight fraction. As shown in \cref{fig:Watson vs P N A}, predicted naphtha via the chemical property-guided model certainly follows general types of compositions behavior of real naphtha in K value increase. Across the full range of Watson K values, the predominant components are paraffin and iso-paraffin, which make up more than 40\% of the total weight. Due to their low boiling points resulting from their straight carbon chain structure, an increase in weight fraction leads to the high Watson K value. Conversely, carbon ring-based aromatic and naphthene components, which have boiling points ranging from 80 to 210$ ^\circ$C, have a relatively lower weight percentage in crude oils with high Watson K values.
Meanwhile, the CNN-based model does not show any composition behavior in different K values. Conclusively, the result reveals that the loss of the Watson K factor guides the training prediction model to generate naphtha composition that reflects realistic chemical properties, rather than just focusing on reducing the prediction accuracy.

\begin{table}[!t]
\caption{\label{table:train_result}Comparison result of the prediction models}
\centering
\begin{tabular}{|c|c|c|} 
 \hline
   Averaged & Naphtha composition & Chemical 
   \\ value & prediction model & property-guided model 
   \\ [0.5ex] 
 \hline
 $MAE$ & $0.0213 \pm 0.005$ & $0.0165 \pm 0.005$ \\ 
  \hline
 $MSE$ & $0.0485 \pm 0.029$ & $0.0327 \pm 0.020$ \\
  \hline
 $R^{2}$ & $0.950 \pm 0.029$ & $0.965 \pm 0.021$ \\
 \hline
\end{tabular}
\end{table}

\section{Conclusion}
\label{sec:conclusion}
The naphtha cracking process is significantly affected by variations in the composition of naphtha, but detailed composition analysis, such as ASTM gas chromatography, is time-consuming and requires pilot-plant scale experiments \cite{bi2019novel}. To address this issue, a neural network framework is proposed in this study, which utilizes chemical property information and laboratory-scaled data to accurately predict complex naphtha composition without detailed experiments. The proposed framework comprises two networks, a Watson K factor estimation network, and a naphtha composition prediction network using distillation curve data. Both networks share a feature extraction network based on the Convolutional Neural Network (CNN) architecture, while the output layers use Multi-Layer Perceptron (MLP) based networks to generate two different outputs: Watson K factor and naphtha composition. The results show that the proposed framework has lower prediction loss than the model using only distillation curve data and shows good agreement with real naphtha data. However, two main limitations remain, 1) the model needs to be evaluated in the cracking process and 2) there is a possibility of overfitting. Although the proposed model has high prediction accuracy, the error in the detailed composition may lead to considerable errors in the cracking process due to its complex reaction kinetics. Therefore, the predicted composition should be tested in the cracking process to ensure that the final product composition is similar to the real composition. Additionally, due to the small dataset size of 254 data points, further studies are required to expand the proposed model for practical application to the petrochemical process. 

\section*{Acknowledgment}

This work was partially supported by Carl-Zeiss Stiftung under the Sustainable Embedded AI project (P2021-02-009) and by the Korean Institute of Industrial Technology under the Development of AI Platform for Continuous Manufacturing of Chemical Process (JH-23-0002) and Development and application of carbon-neutral engineering platform based on carbon emission database and prediction model (KM-23-0098).

\bibliographystyle{IEEEtran}
\bibliography{ref}

\end{document}